\documentclass[10pt,twocolumn,letterpaper]{article}
\usepackage{amsmath}
\usepackage{amssymb}
\usepackage{booktabs}
\usepackage[inline]{enumitem}
\usepackage{epsfig}
\usepackage{graphicx}
\usepackage{iccv}
\usepackage{mathtools}
\usepackage{multirow}
\makeatletter
\@namedef{ver@everyshi.sty}{}
\makeatother
\usepackage{tikz}
\usepackage{pgffor}
\usepackage[font=footnotesize,skip=3pt,subrefformat=parens]{subfig}
\usepackage{tabularx}
\usepackage{times}
\usepackage{xcolor}
\usepackage{subfig}

\graphicspath{{figures/}}

\makeatletter
\renewcommand{\fnum@figure}{Fig. \thefigure}
\makeatother

\newcolumntype{Y}{>{\centering\arraybackslash}X}
\newcolumntype{M}{>{\raggedleft\arraybackslash}m{6mm}}

\def\modelname{LIST}

\PassOptionsToPackage{hyphens}{url}
\usepackage[pagebackref=true,breaklinks=true,letterpaper=true,colorlinks,bookmarks=false]{hyperref}

\iccvfinalcopy 

\graphicspath{{figures/}}

\begin{document}

\title{\modelname: Learning Implicitly from Spatial Transformers\\
for Single-View 3D Reconstruction} 

\author{Mohammad Samiul Arshad and William J. Beksi\\
Department of Computer Science and Engineering\\
The University of Texas at Arlington, Arlington, TX, USA\\
{\tt\small mohammadsamiul.arshad@mavs.uta.edu, william.beksi@uta.edu}
}

\maketitle
\ificcvfinal\thispagestyle{empty}\fi

\begin{abstract}
Accurate reconstruction of both the geometric and topological details of a 3D
object from a single 2D image embodies a fundamental challenge in computer
vision. Existing explicit/implicit solutions to this problem struggle to recover
self-occluded geometry and/or faithfully reconstruct topological shape
structures. To resolve this dilemma, we introduce \modelname, a novel neural
architecture that leverages local and global image features to accurately
reconstruct the geometric and topological structure of a 3D object from a single
image. We utilize global 2D features to predict a coarse shape of the target
object and then use it as a base for higher-resolution reconstruction. By
leveraging both local 2D features from the image and 3D features from the coarse
prediction, we can predict the signed distance between an arbitrary point and
the target surface via an implicit predictor with great accuracy. Furthermore,
our model does not require camera estimation or pixel alignment. It provides an
uninfluenced reconstruction from the input-view direction. Through qualitative
and quantitative analysis, we show the superiority of our model in
reconstructing 3D objects from both synthetic and real-world images against the
state of the art. Our source code is publicly available to the research
community \cite{list}.
\end{abstract}

\section{Introduction}
\label{sec:introduction}
Constructing a truthful portrayal of the 3D world from a single 2D image is a
basic problem for many applications including robot manipulation and
navigation, scene understanding, view synthesis, virtual reality, and more.
Following the work of Erwin Kruppa \cite{kruppa1913ermittlung} in camera motion
estimation and the recovery of 3D points, researchers have attempted to solve
the 3D reconstruction issue using structure from motion
\cite{ullman1979interpretation,longuet1981computer,schonberger2016structure},
and visual simultaneous localization and mapping \cite{fuentes2015visual,
saputra2018visual}. However, the main limitation of such approaches is that
they require multiple observations of the desired object or scene from distinct
viewpoints with shared features. Such a multi-view formulation allows for
integrating information from numerous images to compensate for occluded
geometry.


\begin{figure}[t]
\centering
\includegraphics[width=\columnwidth]{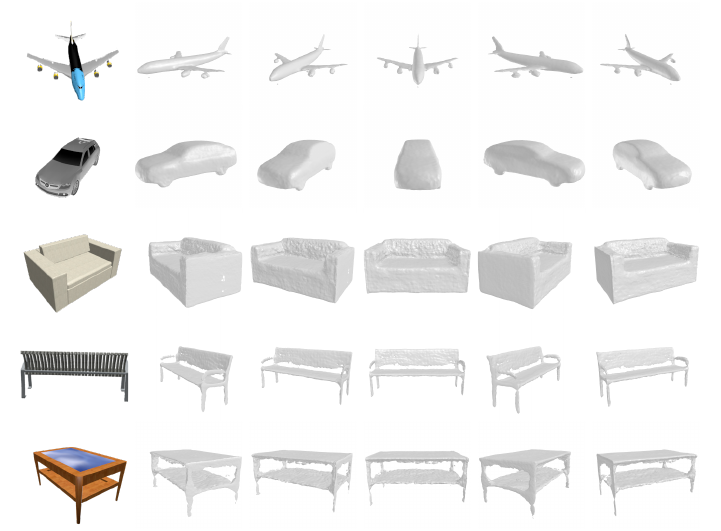}
\caption{Five unique views of objects reconstructed by \modelname~from a single
RGB image. Not only does our model accurately recover occluded geometry, but
also the reconstructed surfaces are \textit{not influenced} by the input-view
direction.}
\label{fig:short_results}
\end{figure}

Reconstructing a 3D object from a single image is a more difficult task since a
sole image does not contain the whole topology of the target shape due to
self-occlusions. Researchers have tried both explicit and implicit techniques
to reconstruct a target object with self-occluded parts. Explicit methods
attempt to infer the target shape directly from the input image. Nevertheless,
a major drawback of such approaches is that the output resolution needs to be
defined in advance, which constrains these techniques from achieving
high-quality results. Recent advances in implicit learning offer a solution to
reconstruct the target shape in an arbitrary resolution by indirectly inferring
the desired surface through a distance/occupancy field. Then, the target
surface is reconstructed by extracting a zero level set from the
distance/occupancy field.

Implicit 3D reconstruction from a single view is an active area of research
where one faction of techniques \cite{mescheder2019occupancy,chen2019learning}
encode global image features into a latent representation and learn an implicit
function to reconstruct the target. Yet, these approaches can be easily
outperformed by simple retrieval baselines \cite{tatarchenko2019single}.
Therefore, global features alone are not sufficient for a faithful
reconstruction. Another faction leverages both local and global features to
learn the target implicit field from pixel-aligned query points. However, such
methods rely on ground-truth/estimated camera parameters for training/inference
\cite{xu2019disn,li2021d2im}, or they assume weak perspective projection
\cite{saito2019pifu,he2020geo}.


To address these shortcomings we propose \modelname, a novel deep learning
framework that can reliably reconstruct the topological and geometric structure
of a 3D object from a single RGB image. Our method \textit{does not depend on
weak perspective projection, nor does it require any camera parameters during
training or inference}. Moreover, we leverage both local and global image
features to generate highly-accurate topological and geometric details. To
recover self-occluded geometry and aid the implicit learning process, we first
predict a coarse shape of the target object from the global image features.
Then, we utilize the local image features and the predicted coarse shape to
learn a signed distance function (SDF).

Due to the scarcity of real-world 2D-3D pairs, we train our model on synthetic
data. However, we use both synthetic and-real world images to test the
reconstruction ability of \modelname. Through qualitative analysis we highlight
our model's \textit{superiority in reconstructing high-fidelity geometric and
topological structure}. Via a quantitative analysis using traditional
evaluation metrics, \textit{we show that the reconstruction quality of
\modelname~surpasses existing works}. Furthermore, \textit{we design a new
metric to investigate the reconstruction quality of self-occluded geometry}.
Finally, we provide an ablation study to validate the design choices of
\modelname~ in achieving high-quality single-view 3D reconstruction.


\section{Related Work}
\label{sec:related_work}
In this section we summarize pertinent work on the reconstruction of 3D objects
from a single RGB image via implicit learning. Interested readers are encouraged
to consult \cite{fu2021single} for a comprehensive survey on 3D reconstruction
from 2D images. Contrary to explicit representations, implicit ones allow for
the recovery of the target shape at an arbitrary resolution. This benefit has
attracted interest among researchers to develop novel implicit techniques for
different applications. Dai \etal \cite{dai2017shape} used a voxel-based
implicit representation for shape completion. DeepSDF, introduced by Park \etal
\cite{park2019deepsdf}, is an auto-decoder that learns to estimate signed
distance fields. However, DeepSDF requires test-time optimization, which limits
its efficiency and capability. 

To further improve 3D object reconstruction quality, Littwin and Wolf
\cite{littwin2019deep} utilized encoded image features as the network weights of
a multilayer perceptron. Wu \etal \cite{wu2020pq} explored sequential part
assembly by predicting the SDFs for structural parts separately and then
combining them together. For self-supervised learning, Liu \etal
\cite{liu2019learning} proposed a ray-based field probing technique to render
the implicit surfaces as 2D silhouettes. Niemeyer \etal
\cite{niemeyer2020differentiable} used supervision from RGB, depth, and normal
images to reconstruct rich geometry and texture. Chen and Zhang
\cite{chen2019learning} proposed generative models for implicit representations
and leveraged global image features for single-view reconstruction. For multiple
3D vision tasks, Mescheder \etal \cite{mescheder2019occupancy} developed OccNet,
a network that learns to predict the probability of a volumetric grid cell being
occupied.

Pixel-aligned approaches \cite{saito2019pifu,saito2020pifuhd,he2020geo,
cao2022jiff} have employed local query feature extraction from image pixels to
improve 3D human reconstruction. Xu \etal \cite{xu2019disn} incorporated similar
ideas for 3D object reconstruction. To enhance the reconstruction quality of
surface details, Li and Zhang \cite{li2021d2im} utilized normal images and a
Laplacian loss in addition to aligned features. Zhao \etal
\cite{zhao2021learning} exploited coarse prediction and unsigned distance fields
to reconstruct garments from a single view. Duggal and Pathak
\cite{duggal2022topologically} proposed category specific reconstruction by
learning a topology aware deformation field. Mittal $\etal$ introduced AutoSDF
\cite{mittal2022autosdf}, a model that encodes local shape regions separately
via patch-wise encoding. However, these prior works rely on weak perspective
projection and the rendering of metadata to align query points to image pixels.
In contrast, \modelname~does not require any alignment or rendering data, and it
recovers more accurate topological structure and geometric details.

\section{Implicit Function Learning from Unaligned Pixel Features} 
\label{sec:method}
\begin{figure*}
\centering
\includegraphics[width=0.9\textwidth, height=0.3\textheight]{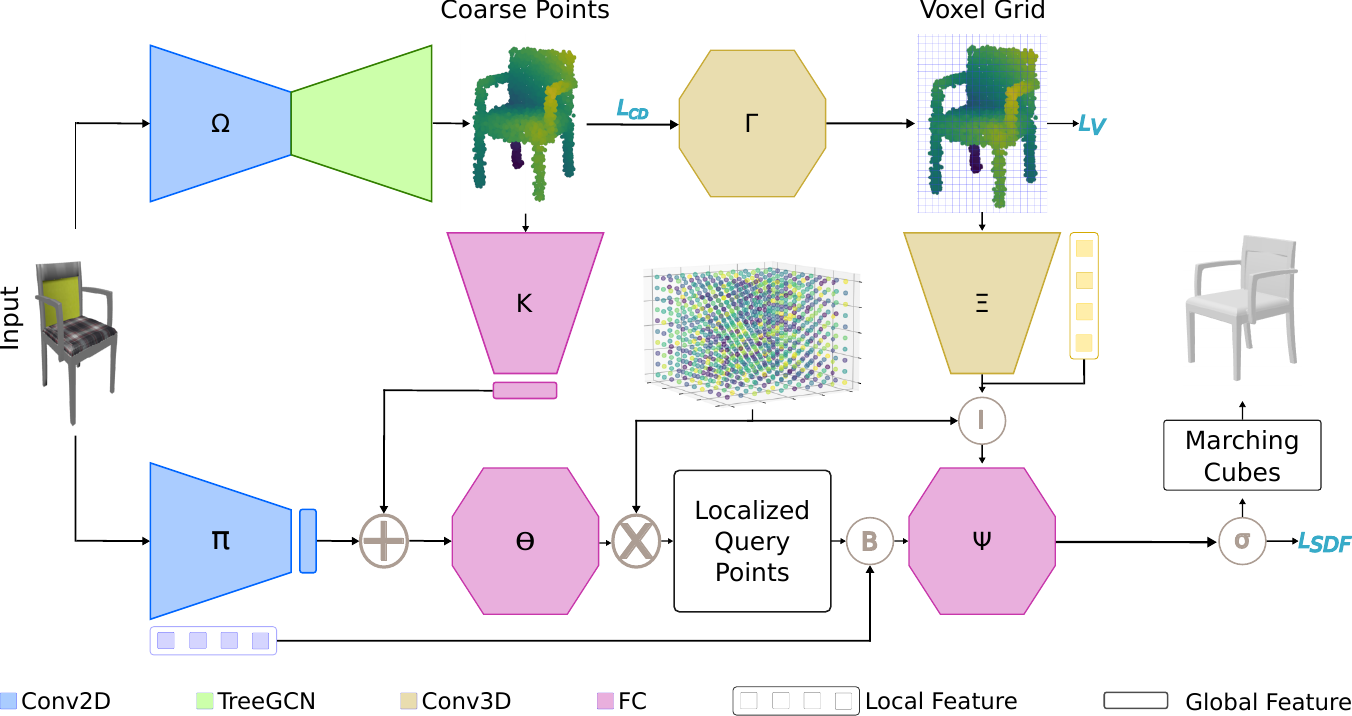}
\caption{To reconstruct the target object from a single RGB image,
\modelname~first predicts the coarse topology from the global image features.
Simultaneously, local image features are used to extract local geometry at the
given query locations. Finally, an SDF predictor ($\Psi$) estimates the signed
distance field ($\sigma$) to reconstruct the target shape. Note that images and
colors are for visualization purposes only.}
\label{fig:model}
\end{figure*}

Given a single RGB image of an object, our goal is to reconstruct the object in
3D with highly-accurate topological structure and self-occluded geometry. We
model the target shape as an SDF and extract the underlying surface from the
zero level set of the SDF during inference. To train our model we employ an
image and query point pair ($x_i,Q_i$), where $Q_i$ is a set of 3D coordinates
(query points) in close vicinity to the surface of the object with a measured
signed distance and $x_i$ is a rendering of the object from a random viewpoint.
An overview of the our framework is presented in Fig.~\ref{fig:model}. The
details of each component are provided in the following subsections.


\subsection{Query Features From Coarse Predictions}
Consider an RGB image $x_i \subset X \in \mathbb{R}^{H \times W \times 3}$ of
height $H$ and width $W$. We propose a convolutional neural encoder-decoder
$\Omega_\omega$, parameterized by weights $\omega$, to extract latent features
from the image and predict a coarse estimation $\dot{y}_i^{x_i}$
of the target object. Concretely,
\begin{equation}
  \Omega_\omega(x_i) \coloneqq \dot{y}_i^{x_i} ~|~ \mathbb{R}^{H \times W \times 3} \rightarrow \mathbb{R}^{N \times 3},
\end{equation}
where $\dot{y}_i^{x_i}$ is a point cloud representation of the target and $N$ is
the resolution of the point cloud. Note that the subscript $i$ indicates $i$-th
sample and the superscript $x_i$ designates the source variable. For
high-performance point cloud generation, we utilize tree structured graph
convolutions (TreeGCN) \cite{shu20193d} to decode the image features.

We use the coarse prediction $\dot{y}_i$ as a guideline for the topological
structure of the target shape in a canonical space. To extract query features
from this coarse prediction, first we discretize the point cloud in an occupancy
grid $\dot{u}_i^{\dot{y}_i} \in 1^{M \times M \times M}$ of resolution $M$.
However, the coarse prediction may contain gaps and noisy points that may impair
the reconstruction quality. To resolve this, we employ a shallow convolutional
network $\Gamma_{\ddot{o}}$ parameterized by weights $\ddot{o}$ to generate a
probabilistic occupancy grid from $\dot{u}_i^{\dot{y}_i}$,
\begin{equation}
  \dot{v}_i^{\dot{u}_i} \coloneqq \Gamma_{\ddot{o}}(\dot{u}_i^{\dot{y}_i}) \colon 1^{M \times M \times M} \rightarrow [0,1]^{M \times M \times M}.
  \label{eq:occupancy_grid_estimation}
\end{equation}
Specifically, our aim is to find the neighboring points of $\dot{y}_i$ with a
high chance of being a surface point of the target shape.

Although it is possible to regress the voxel representation directly from the
global image features \cite{choy20163d,sitzmann2019deepvoxels,he2020geo},
learning a high-resolution voxel occupancy prediction requires a
\textit{significant} amount of computational resources \cite{he2020geo}.
Moreover, we empirically found that point cloud prediction followed by voxel
discretization achieves better accuracy on diverse shapes rather than predicting
the voxels directly.

Next, a neural network $\Xi_\xi$, parameterized by weights $\xi$, maps the
probabilistic occupancy grid \eqref{eq:occupancy_grid_estimation} to a
high-dimensional latent matrix through convolutional operations. Then, our
multi-scale trilinear interpolation scheme $\mathit{I}$ extracts relevant query
features $f_{C}$ at each query location $q_i$ from the mapped features. More
formally,
\begin{equation}
  f_{C} \coloneqq \mathit{I}(\Xi_\xi(\dot{v}_i^{\dot{u}_i}), Q_i).
\end{equation}
In addition to $q_i$, we also consider the neighboring points at a distance $d$
from $q_i$ along the Cartesian axes to capture rich 3D features,
i.e.,
\begin{equation}
  q_j = q_j + k \cdot \hat{n}_j \cdot d,
\end{equation}
where $k \in \{1, 0, -1\}$, $j \in \{1, 2, 3\}$, and $\hat{n}_j \in
\mathbb{R}^3$ is the $j$-th Cartesian axis unit vector.

\subsection{Localized Query Features}
The coarse prediction and query features $f_{C}$ can aid the recovery of the
topological structure of the target shape. Nevertheless, relevant local features
are also required to recover fine geometric details. To achieve this, prior arts
assume weak perspective projection \cite{saito2019pifu,he2020geo} or align the
query points to the image pixel locations through the ground-truth/estimated
camera parameters \cite{xu2019disn,li2021d2im}. Predicting the camera parameters
is analogous to predicting the object pose from a single image, which is itself
a hard problem in computer vision. It involves a high chance of error and a
computationally expensive training procedure. Furthermore, the error in the
pose/camera estimation may lead to the loss of geometric details in the
reconstruction.

To overcome these limitations, we obtain insight from spatial transformers
\cite{jaderberg2015spatial} and leverage the spatial relationship between the
input image and the coarse prediction. Via the coarse prediction, which portrays
an object from a standard viewpoint and the query points that delineate the
coarse predictions, it is possible to localize the query points to the local
image features. This is done by predicting a spatial transformation with the aid
of global features from the input image and the coarse prediction as follows.

First, we define a convolutional neural encoder $\Pi_\pi$, parameterized by
weights $\pi$, to encode the input image into local $(l^{x_i}_\pi)$ and global
$(z^{x_i}_\pi)$ features. Concretely,
\begin{equation}
  \Pi_\pi(x_i) \coloneqq \{l^{x_i}_\pi, z^{x_i}_\pi\}.
\end{equation}
Concurrently, a neural module $K_\kappa$ encodes the coarse prediction
$\dot{y}_i^{x_i}$ into global point features. Using global features from both
the image and the coarse prediction, the spatial transformer $\Theta$ estimates
a transformation to localize the query points in the image feature space. Then,
localized query points $\tilde{Q_i}$ are generated by applying the predicted
transformation to $Q_i$,
\begin{equation}
  \Theta_\theta(z_\pi^{x_i}, K_\kappa(\dot{y}_i^{x_i}), Q_i) \coloneqq \tilde{Q_i} ~|~ \mathbb{R}^{N \times 3} \rightarrow \mathbb{R}^{N \times 2}.
\end{equation} 
Finally, a bi-linear interpolation scheme $\mathcal{B}$ extracts the local query
features $f_L$ from the local image features $l^{x_i}_\pi$,
\begin{equation}
  \mathit{f}_{L} \coloneqq \mathcal{B}(l_\pi^{x_i}, \tilde{Q_i}).
\end{equation}

Note that the point encoder $K_\kappa$ and the localization network $\Theta$ are
designated to ensure an accurate SDF prediction. Therefore, we do not use any
camera parameters during training and we optimize these neural modules directly
with the SDF prediction objective. This has the following benefits:
\begin{enumerate*}[label=(\roman*)]
  \item \textit{additional modules or training to predict the projection matrix
  and object pose from a single image are not required;} 
  \item \textit{reconstructions are free from any pose estimation error, which
  boosts reconstruction accuracy}.
\end{enumerate*}

\subsection{Signed Distance Function Prediction}
To estimate the final signed distance $\Delta_i$, we combine the coarse features
$f_C$ with the localized query features $f_L$ and utilize a multilayer neural
function defined as 
\begin{equation}
  \Psi_\psi(f_C, f_L) \coloneqq 
  \begin{cases}
    \mathbb{R}^{-}, & \text{if $q_i$ is inside the target surface}\\
    \mathbb{R}^{+}, & \text{otherwise}.
\end{cases}
\end{equation}

\subsection{Loss Functions}
We incorporate the chamfer distance (CD) loss and optimize the weights $\omega$
to accurately estimate the coarse shape of the target. More specifically,
\begin{equation}
  \mathcal{L}_\text{CD}(y_i, \dot{y}_i) = \sum_{a\in \dot{y}_i}^{}\min_{b \in y_i} {||a-b||}^2 + \sum_{b\in y_i}^{}\min_{b\in \dot{y}_i} {||b-a||}^2,
\end{equation}
where $y_i \in \mathbb{R}^{N \times 3}$ is a set of 3D coordinates collected
from the surface of the object and $\dot{y}_i \in \mathbb{R}^{N \times 3}$ is
the estimated coarse shape. To supervise the probabilistic occupancy grid
prediction, we discretize $y_i$ to generate the ground-truth occupancy
$v_{i}^{y_i}\in 1^{M \times M \times M}$.
The neural weight $\ddot{o}$ is then optimized by the binary cross-entropy loss,
\begin{equation}
  \mathcal{L}_V(v_i, \dot{v}_i) = -\frac{1}{|v_i|}\Sigma(\gamma v_i\log{\dot{v}_i} + (1-\gamma) (1-v_i)\log(1-{\dot{v}_i})),
\end{equation}
where $\gamma$ is a hyperparameter to control the influence of the
occupied/non-occupied grid points. To optimize the SDF prediction, we collect a
set of query points $Q_i$ within distance $\delta$ of the target surface and
measure their signed distance $\sigma_i$. The estimated signed distance is then
guided by optimizing the neural weights $\xi$, $\pi$, $\theta$, and $\psi$
through
\begin{equation}
  \mathcal{L}_{\text{SDF}} = \frac{1}{|Q_i|}\Sigma(\sigma_i - \Delta_i)^2.
\end{equation}

\subsection{Training Details}
We incorporate a two-stage procedure to train \modelname. In the first stage, we
only focus on the coarse prediction from the input image $x_i$ and optimize the
weights $\omega$ through $\mathcal{L}_\text{CD}$. Then, we freeze $\omega$ after
convergence to a minimum validation accuracy and start the second stage for the
SDF prediction. During the second stage, we jointly optimize $\ddot{o}$, $\xi$,
$\pi$, $\kappa$, $\theta$, and $\psi$ through the combined loss $\mathcal{L} =
\mathcal{L}_V + \mathcal{L}_{\text{SDF}}$. \modelname\ can also be trained
end-to-end by jointly minimizing $\mathcal{L}_{CD}$ with $\mathcal{L}_V$ and
$\mathcal{L}_{\text{SDF}}$. However, we found the two-stage training procedure
easier to evaluate and quicker to converge during experimental evaluation. To
reconstruct an object at test time, we first densely sample a fixed 3D grid of
query points and predict the signed distance for each point.  Then, we use the
marching cubes \cite{lorensen1987marching} algorithm to extract the target
surface from the grid.

\section{Experimental Evaluation}
\label{sec:experimental_evaluation}
In this section, we describe the details of our experimental setup and results.
Additional information, including implementation details, can be found in the
supplementary material.

\subsection{Datasets}
Similar to \cite{li2021d2im} and \cite{mittal2022autosdf}, we utilized the
13-class subset of the ShapeNet \cite{chang2015shapenet} dataset to train
\modelname. The renderings and processed meshes from \cite{xu2019disn} were used
as the input view and target shape. We trained a single model on all 13
categories. Additionally, we employed the Pix3D \cite{sun2018pix3d} dataset to
test \modelname~on real-world scenarios. The train/test split from
\cite{zhang2021holistic} was used to evaluate on all 9 categories of Pix3D.
Following \cite{zhang2021holistic}, we preprocessed the Pix3D target shapes to
be watertight for training.

To prepare the ground-truth data, we first normalized the meshes to a unit cube
and then sampled 50 k points from the surface of each object. Next, we
displaced the sampled points with a Normal distribution of zero mean and
varying standard deviation. Lastly, we calculated the signed distance for every
point. To supervise the coarse prediction and probabilistic occupancy grid
estimation, we sub-sampled 4 k points from the surface via farthest point
sampling. Further details regarding the data preparation strategy can be found
in the supplementary material.

\subsection{Baseline Models}
For single-view reconstruction via synthetic images, we compared against the
following prior arts: IMNET \cite{chen2019learning}, and D$^2$IM-Net
\cite{li2021d2im}. IMNET does not require pose estimation. However, the
reconstruction only unitizes global features from an image. D$^2$IM-Net extracts
local features by aligning the query points to image pixels through rendering
metadata and it uses a pose estimation module during inference.

For single-view reconstruction from real-world images, we evaluated against TMN
\cite{pan2019deep}, MGN \cite{nie2020total3dunderstanding}, and IM3D
\cite{zhang2021holistic}. TMN deforms a template mesh to reconstruct the target
object. MGN and IM3D perform reconstruction through the following steps: (i)
identify objects in a scene, (ii) estimate their poses, and (iii) reconstruct
each object separately. 


\subsection{Metrics}
\begin{figure}
\centering
\subfloat[]{\includegraphics[trim={50 50 50 50},clip,
width=0.2\columnwidth]{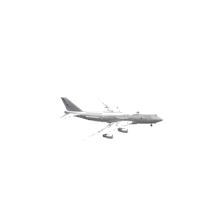}}
\subfloat[]{\includegraphics[trim={50 50 50 50},clip,
width=0.2\columnwidth]{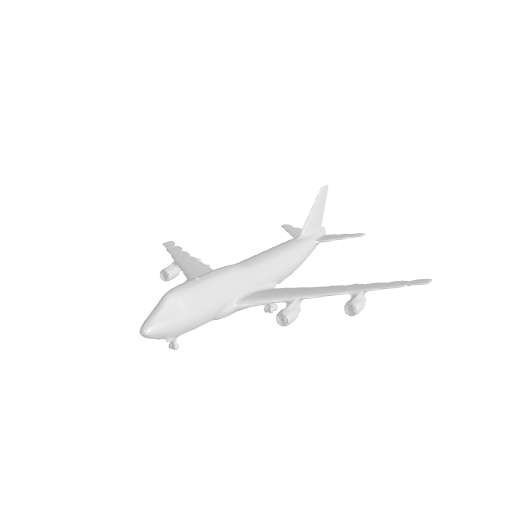}}
\subfloat[]{\includegraphics[trim={50 50 50 50},clip,
width=0.2\columnwidth]{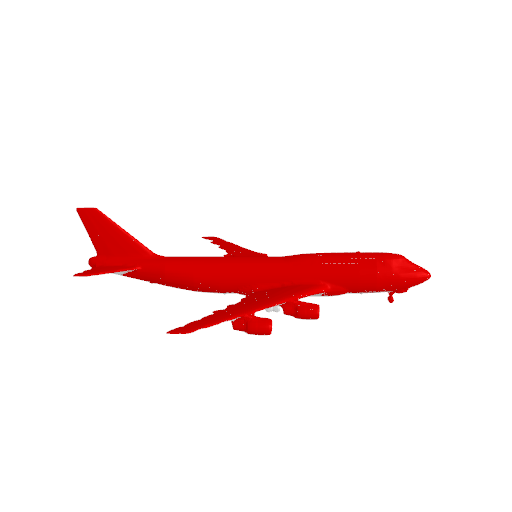}}
\subfloat[]{\includegraphics[trim={50 50 50 50},clip,
width=0.2\columnwidth]{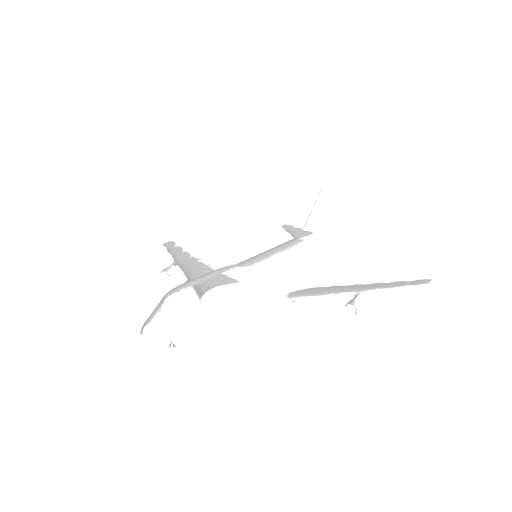}}
\subfloat[]{\includegraphics[trim={50 50 50 50},clip,
width=0.2\columnwidth]{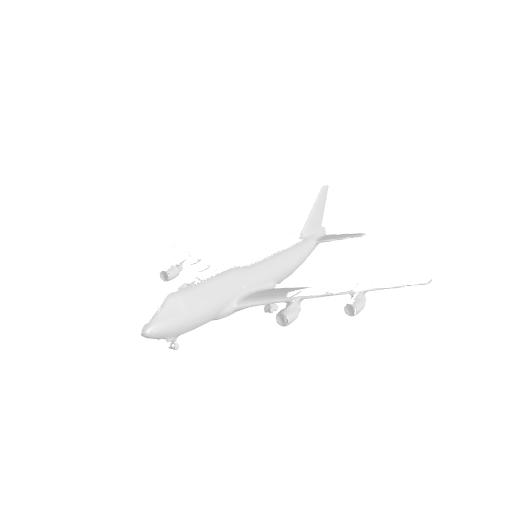}}
\caption{To evaluate the reconstruction quality of occluded surfaces, we first
align the reconstructed shape (b) with the input image (a) and cast rays onto
the surface (c). Next, we identify the (red) faces that intersect with the rays
via ray-mesh intersection and separate the reconstructed mesh into (d) visible
and (e) occluded areas.}
\label{fig:occluded_evaluation}
\end{figure}

We computed commonly used metrics (e.g., CD, intersection over union (IoU), and
F-score), to evaluate the performance of \modelname. The definitions of these
metrics can be found in the supplementary material. Nonetheless, these
traditional metrics \textit{do not} differentiate between visible/occluded
surfaces since they evaluate the reconstruction as a whole. To investigate the
reconstruction quality of occluded surfaces, we propose to isolate
visible/occluded surfaces based on the viewpoint of the camera and evaluate them
separately using the traditional metrics. A visual depiction of this new
strategy is presented in Fig.~\ref{fig:occluded_evaluation}.

To measure the reconstruction quality of occluded surfaces, we first align the
predicted/ground-truth meshes to their projection in the input image using the
rendering metadata. Then, we assume the camera location as a single source of
light and cast rays onto the mesh surface by ray casting \cite{roth1982ray}.
Next, we identify the visible/occluded faces through the ray-mesh intersection
and subdivide the identified faces to separate them. Note that the rendering
metadata is only used to evaluate the predictions. Finally, we sample 100 k
points from the separated occluded faces to compute the $\text{CD}_\text{os}$,
and voxelize the sampled points to compute the $\text{IoU}_\text{os}$ and
$\text{F-Score}_\text{os}$.

In our implementation, we set the canvas resolution to $4096\times 4096$ pixels
and generated one ray per pixel from the camera location. It is important to
note that ray casting and computing ray-mesh intersections are computationally
demanding tasks. Therefore, to manage time and resources, we chose five
sub-classes (chair, car, plane, sofa, table) to evaluate occluded surface
reconstruction. 

\subsection{Single-View 3D Reconstruction Evaluation}
\begin{figure*}
\centering
\includegraphics[width=0.95\textwidth, trim=0 0 30 0, clip]{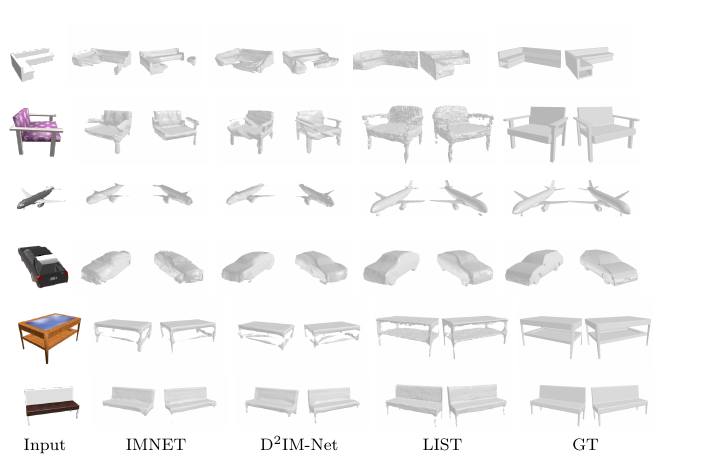}
\caption{A qualitative comparison between \modelname~and the baseline models
using the ShapeNet \cite{chang2015shapenet} dataset. Our model recovers
\textit{significantly better} topological and geometric structure, and the
reconstruction is not tainted by the input-view direction. GT denotes the
ground-truth objects.}
\label{fig:qualitative_results_shapenet}
\end{figure*}

\setlength\tabcolsep{0.87mm}
\begin{table*}
\begin{tabularx}{\textwidth}{c|c|ccccccccccccc|c} \hline
&       & plane           & bench           & cabinet         & car
        & chair           & display         & lamp            & speaker
        & rifle           & sofa            & table           & phone    
        & boat            & Mean \\ \hline
\multirow{4}{*}{CD$\downarrow$} 
& IMNET 
        & 18.95           & 17.34           & 15.17           & 10.86
        & 14.72           & 16.77           & 83.64           & 33.41    
        & 10.33           & 13.35           & 19.32           & 9.16
        & 15.24           & 21.40 \\
& D$^2$IM-Net
        & 13.25           & \textbf{12.51}  & 9.47            & 7.83
        & 11.31           & 15.33           & \textbf{34.08}           & 17.62
        & 8.55            & 12.34           & 14.26           & 8.11      
        & 15.73           & 13.87 \\
        
& \modelname
        & \textbf{12.13}  & 13.49           & \textbf{7.45}   & \textbf{1.04}
        & \textbf{9.20}   & \textbf{13.65}  & 47.31           & \textbf{16.75}
        & \textbf{7.32}   & \textbf{9.92}   & \textbf{11.14}  & \textbf{7.91}
        & \textbf{15.78}  & \textbf{13.31} \\ \hline
\multirow{4}{*}{IoU$\uparrow$} 
& IMNET 
        & 39.43           & 44.65           & 49.25           & 55.75
        & 51.22           & 53.34           & 29.26           & 50.66
        & 46.43           & 51.12           & 41.63           & 52.79
        & 49.61           & 47.31\\
& D$^2$IM-Net
        & 45.44           & 48.45           & 48.60            & 53.58      
        & 53.13           & 52.72           & \textbf{32.45}   & 51.75
        & 50.76           & 53.35           & 45.17            & 53.06            & 
        52.89             & 49.33 \\
& \modelname
        & \textbf{49.03}  & 47.57           & \textbf{56.29}   & \textbf{65.57}
        & \textbf{52.70}  & \textbf{57.34}  & 24.80            & \textbf{55.34}
        & \textbf{52.42}  & \textbf{56.79}  & \textbf{47.90}   & \textbf{58.98}
        & \textbf{54.35}  & \textbf{52.23} \\ \hline
\multirow{4}{*}{F-score$\uparrow$} 
& IMNET 
        & 48.87           & 31.78           & \textbf{44.34}  & 48.78
        & 41.45           & 48.32           & 21.23           & 48.29   
        & 52.92           & 44.12           & 45.21           & 51.52   
        & 52.31           & 44.54 \\
& D$^2$IM-Net
        & 51.37           & \textbf{36.76}  & 43.49           & 51.77  
        & 45.56           & 50.82           & \textbf{29.57}  & 51.93  
        & 56.25           & 48.34           & 47.23           & 54.84   
        & 52.73           & 47.74 \\
& \modelname
        & \textbf{52.46}  & 36.39           & 42.51           & \textbf{53.12}
        & \textbf{46.62}  & \textbf{51.78}  & 22.88           & \textbf{52.67}
        & \textbf{58.24}  & \textbf{50.52}  & \textbf{49.62}  & \textbf{56.89}
        & \textbf{53.58}  & \textbf{48.25} \\ \hline
\end{tabularx}
\caption{Quantitative results using the ShapeNet \cite{chang2015shapenet}
dataset for various models. The metrics reported are the following: chamfer
distance (CD), intersection over union (IoU), and F-score. The CD values are
scaled by $10^{-3}$.} 
\label{tab:shapenet_results}
\end{table*}

\begin{figure}
\centering
\includegraphics[width=\columnwidth, height=7cm, trim=0 0 50 0, clip]{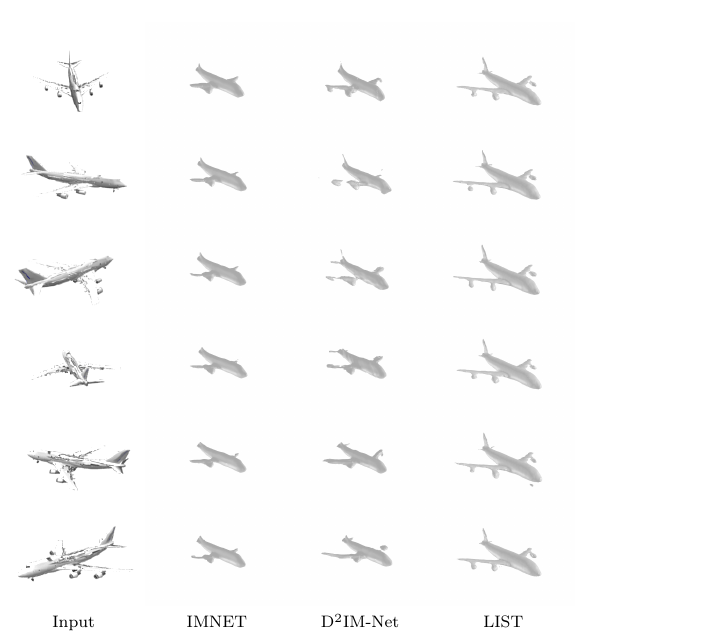}
\caption{A qualitative comparison between \modelname~and the baseline models
using distinct views of the same object. Not only can our model both maintain
better topological structure and geometric details, but it also provides a
reconstruction that is stable across different views of the object.}
\label{fig:qualitative_results_different_views}
\end{figure}

\setlength\tabcolsep{0.45mm}
\begin{table}
\begin{tabularx}{\columnwidth}{c|c|ccccc|c} \hline
&       & plane           & car             & chair           & sofa
        & table           & Mean \\ \hline
\multirow{4}{*}{CD$_\text{os}\downarrow$} 
& IMNET 
        & 24.11           & 13.34           & 15.47           & 24.34
        & 26.86           & 20.82           \\
& D$^2$IM-Net
        & 26.23           & 13.44           & 13.59           & 20.45
        & 23.45           & 19.43           \\
& \modelname
        & \textbf{18.93}  & \textbf{6.57}   & \textbf{12.66}  & \textbf{18.44}
        & \textbf{21.76}   & \textbf{15.67}  \\ \hline
\multirow{4}{*}{IoU$_\text{os}\uparrow$} 
& IMNET 
        & 45.63           & 46.87           & 38.32           & 45.87
        & 39.02           & 43.14           \\
& D$^2$IM-Net
        & 48.44           & 50.33           & 49.43           & 50.32      
        & 42.22           & 48.14          \\
& \modelname
        & \textbf{53.15}  & \textbf{55.37}  & \textbf{51.25}  & \textbf{55.22}
        & \textbf{43.17}  & \textbf{51.63}  \\ \hline
\multirow{4}{*}{F$_\text{os}$-score$\uparrow$} 
& IMNET 
        & 40.93           & 46.94           & 44.43           & 46.84
        & 45.64           & 44.95           \\
& D$^2$IM-Net
        & 47.21           & 50.73           & 48.89           & 49.15  
        & 47.72           & 48.73          \\
& \modelname
        & \textbf{50.33}  & \textbf{52.55}  & \textbf{49.34}  & \textbf{51.02}
        & \textbf{48.11}  & \textbf{50.27}  \\ \hline
\end{tabularx}
\caption{A quantitative evaluation of the occluded surfaces of reconstructed
synthetic objects via our evaluation strategy. The metrics reported are the
following: chamfer distance (CD$_\text{os}$), intersection over union
(IoU$_\text{os}$), and F$_\text{os}$-score. The CD$_\text{os}$ values are scaled
by $10^{-3}$.}
\label{tab:occlusion_evaluation_results}
\end{table}

\subsubsection{Single-View 3D Reconstruction from Renderings of Synthetic
Objects}
In this experiment we performed single-view 3D reconstruction on the test set of
the ShapeNet dataset. The qualitative and quantitative results are displayed in
Fig.~\ref{fig:qualitative_results_shapenet} and
Table~\ref{tab:shapenet_results}, respectively. In comparison to the baselines,
the topological structure and occluded geometry recovered by \modelname~are
considerably better. For example, in row 3 all of the baselines struggle to
reconstruct the tail of the airplane and they fail to estimate the full length
of the wings. In row 5, none of the baselines were able to recover the occluded
part of the table. In contrast, \modelname~not only recovers the structure, but
it also maintains the gap in between. Moreover, notice that in row 2 D$^2$IM-Net
fails to resolve the directional view ambiguity and imprints an arm shaped
silhouette on the seat rather than reconstructing the arm. This indicates a
strong influence of the input-view direction in the reconstructed surface.
Conversely, \modelname~can resolve view-directional ambiguity and provide a
reconstruction that is uninfluenced by the input-view direction. As shown in
Table~\ref{tab:shapenet_results}, \modelname~outperforms all the other baseline
models.

We also evaluated \modelname~against the baselines on occluded surface recovery
by partitioning the reconstructions using our proposed metric. The results are
recorded in Table~\ref{tab:occlusion_evaluation_results}.
\modelname~outperformed all the baselines hence showcasing the superiority of
our approach in reconstructing occluded geometry. Furthermore,
\modelname~provides a stable reconstruction across different views of the same
object as shown in Fig.~\ref{fig:qualitative_results_different_views}. However,
the use of ground-truth rendering data instead of the estimated data improved
the reconstruction quality. This indicates the source of the problem to be the
sub-optimal prediction of the camera pose. Nonetheless, \modelname~is free from
any such complication as our framework does not require any explicit pose
estimation.

\subsubsection{Single-View 3D Reconstruction from Real Images}

\setlength\tabcolsep{2.7mm}
\begin{table*}
\begin{tabularx}{\textwidth}{c|c|ccccccccc|c} \hline
&       & bed             & bookcase        & chair           & desk
        & sofa            & table           & tool            & wardrobe
        & misc            & Mean \\ \hline
\multirow{4}{*}{CD$\downarrow$} 
& TMN 
        & 7.78            & 5.93            & 6.86            & 7.08
        & 4.25            & 17.42           & 4.13            & 4.09
        & 23.68           & 9.03 \\
& MGN  
        & 5.99            & 6.56            & \textbf{5.32}   & 5.93
        & 3.36            & 14.19           & 3.12            & 3.83
        & 26.93           & 8.36 \\
& IM3D
        & \textbf{4.11}   & 3.96            & 5.45            & 7.85
        & 5.61            & 11.73           & 2.39            & 4.31
        & 24.65           & 6.72 \\
& \modelname
        & 5.81            & \textbf{1.74}   & 6.11            & \textbf{3.87}
        & \textbf{2.08}   & \textbf{1.68}   & \textbf{1.99}   & \textbf{0.80}
        & \textbf{5.16}   & \textbf{4.36} \\ \hline
\multirow{1}{*}{IoU$\uparrow$} 
& \modelname
        & 45.61           & 39.54           & 41.15           & 59.68
        & 67.34           & 49.12           & 27.82           & 43.87
        & 34.72           & 46.77 \\ \hline
\multirow{1}{*}{F-score$\uparrow$} 
& \modelname
        & 58.18           & 67.22           & 60.01           & 78.34
        & 70.14           & 69.19           & 46.48           & 75.70
        & 39.14           & 65.66 \\
\end{tabularx}
\caption{A quantitative evaluation of the occluded surfaces of reconstructed
real-world objects using our evaluation strategy. The metrics reported are the
following: chamfer distance (CD$_\text{os}$), intersection over union
(IoU$_\text{os}$), and F$_\text{os}$-score. The CD$_\text{os}$ values are scaled
by $10^{-3}$.}
\label{tab:pix3d_evaluation_results}
\end{table*}

\captionsetup[subfigure]{labelformat=empty,font=footnotesize}
\begin{figure}
\centering
\subfloat{
  \includegraphics[width=0.19\columnwidth]{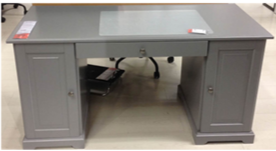}
}
\subfloat{
  \includegraphics[width=0.19\columnwidth, trim={0 10 0 0}, clip]
  {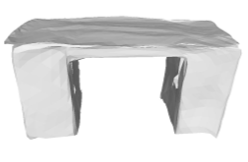}
}
\subfloat{
  \includegraphics[width=0.19\columnwidth, trim={0 10 0 0}, clip]
  {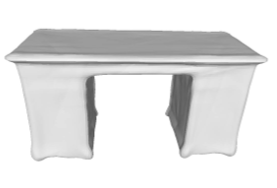}
}
\subfloat{
  \includegraphics[width=0.19\columnwidth, trim={20 120 20 0}, clip]
  {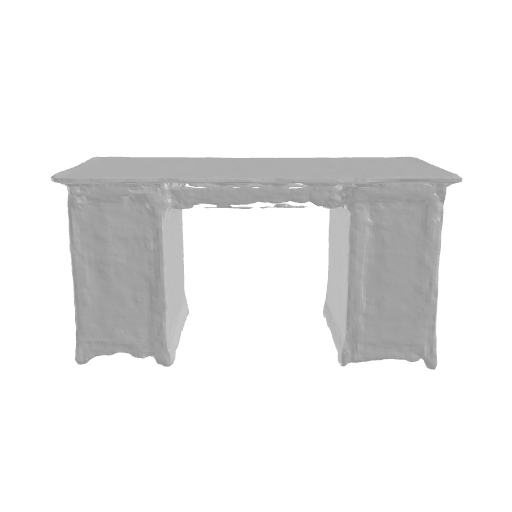}
}
\subfloat{
  \includegraphics[width=0.19\columnwidth, trim={20 120 20 0}, clip]
  {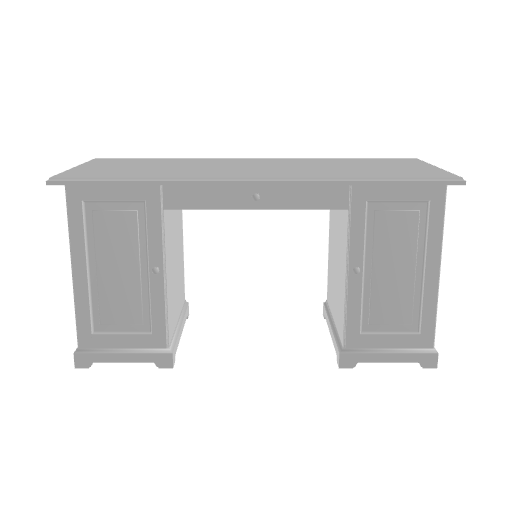}
} \\
\vspace{-0.75em}
\subfloat{
  \includegraphics[width=0.19\columnwidth]{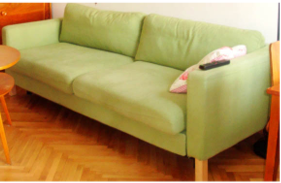}
}
\subfloat{
  \includegraphics[width=0.19\columnwidth, trim={0 10 0 0}, clip]
  {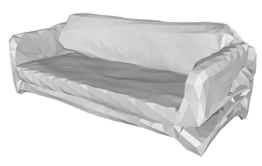}
}
\subfloat{
  \includegraphics[width=0.19\columnwidth, trim={0 10 0 0}, clip]
  {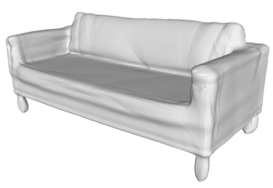}
}
\subfloat{
  \includegraphics[width=0.19\columnwidth, trim={50 120 50 120}, clip]
  {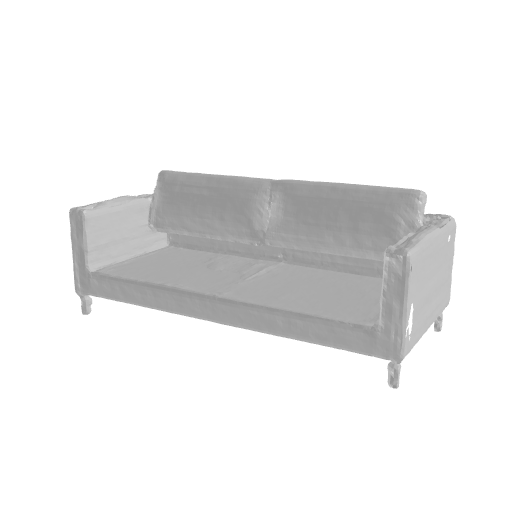}
}
\subfloat{
  \includegraphics[width=0.19\columnwidth, trim={50 120 50 120}, clip]
  {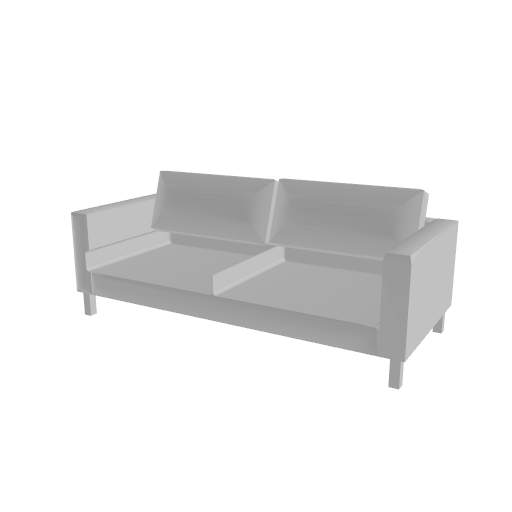}
} \\
\vspace{-1.5em}
\subfloat{
  \includegraphics[width=0.19\columnwidth, height=2.2cm]
  {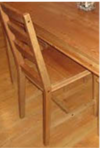}
}
\subfloat{
  \includegraphics[width=0.19\columnwidth, trim={0 0 0 0}, clip]
  {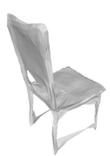}
}
\subfloat{
  \includegraphics[width=0.19\columnwidth, trim={0 0 0 0}, clip]
  {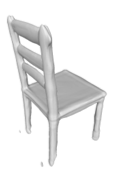}
}
\subfloat{
  \includegraphics[width=0.19\columnwidth, trim={100 30 100 0}, clip]
  {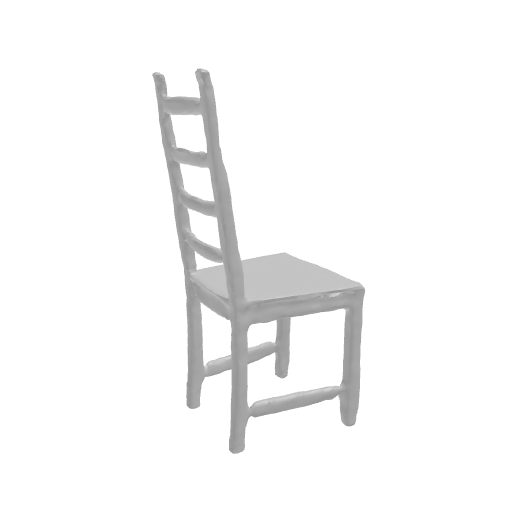}
}
\subfloat{
  \includegraphics[width=0.19\columnwidth, trim={100 30 100 0}, clip]
  {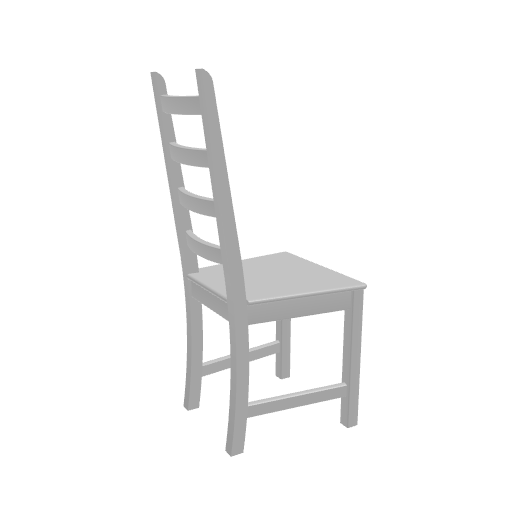}
} \\
\vspace{-1.5em}
\subfloat[Input]{
  \includegraphics[width=0.19\columnwidth, height=2.2cm]{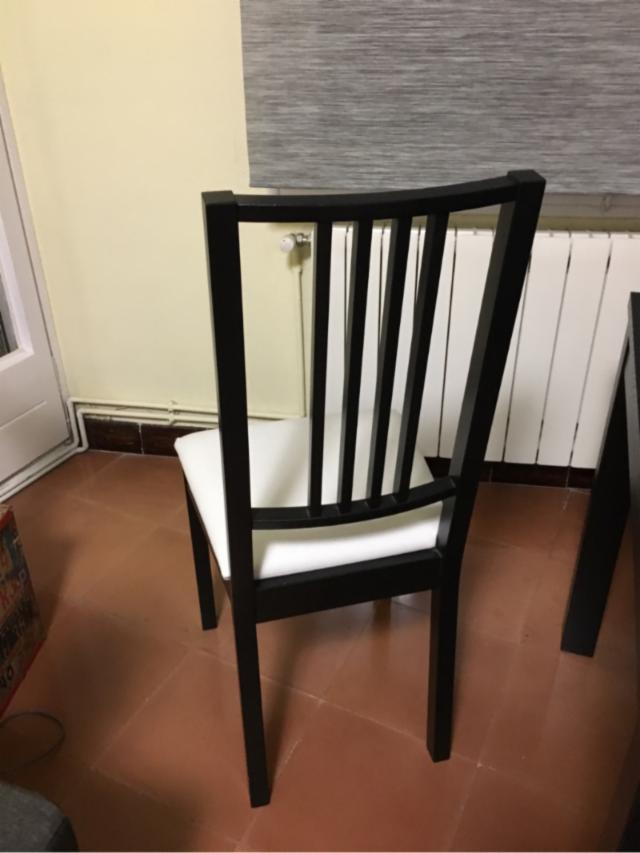}
}
\subfloat[MGN]{
  \includegraphics[width=0.19\columnwidth, trim={0 0 0 0}, clip, height=2.5cm]
  {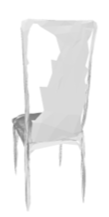}
}
\subfloat[IM3D]{
  \includegraphics[width=0.19\columnwidth, trim={0 0 0 0}, clip, height=2.5cm]
  {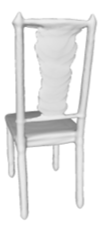}
}
\subfloat[\modelname]{
  \includegraphics[width=0.19\columnwidth, trim={120 30 120 20}, clip, height=2.5cm]
  {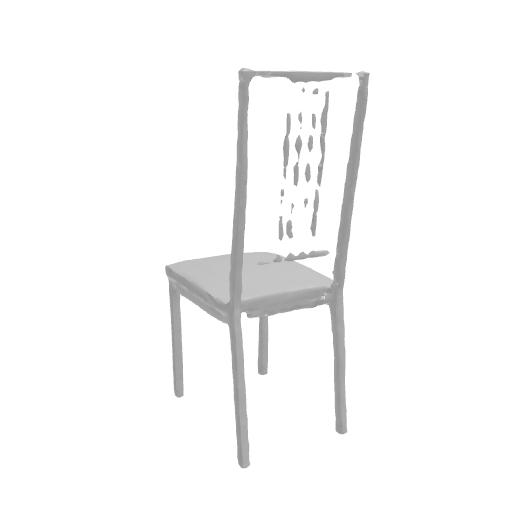}
}
\subfloat[GT]{
  \includegraphics[width=0.19\columnwidth, trim={120 30 120 20}, clip, height=2.5cm]
  {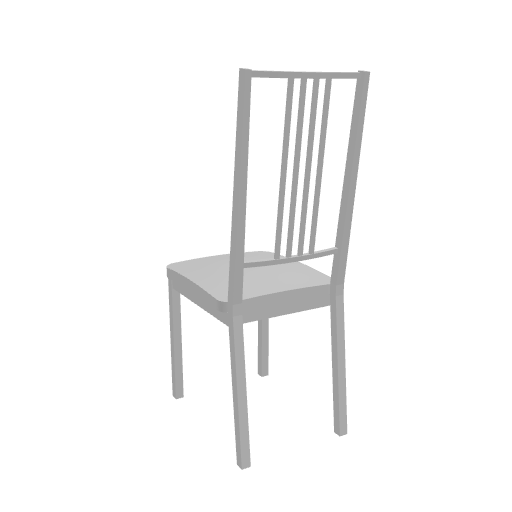}
} \\
\caption{Single-view reconstruction using real-world images from the Pix3D
\cite{sun2018pix3d} test set (best viewed zoomed in).}
\label{fig:qualitative_results_real}
\end{figure}

In this experiment we evaluated single-view 3D reconstruction on the test set of
the Pix3D dataset. The qualitative and quantitative results are provided in
Fig.~\ref{fig:qualitative_results_real} and
Table~\ref{tab:pix3d_evaluation_results}, respectively. The baseline results
were obtained from the respective papers. Compared to other methods our approach
generates the most precise 3D shapes, which results in the lowest average CD and
F-score. Notice that in Fig.~\ref{fig:qualitative_results_real}, rows 3 and 4,
only \modelname~can accurately recover the back and legs of the chair.
Additionally, \modelname~reconstructions provide a smooth surface, precise
topology, and fine geometric details.

\subsection{Ablation Study}
\begin{figure}
\centering
\includegraphics[width=\columnwidth]{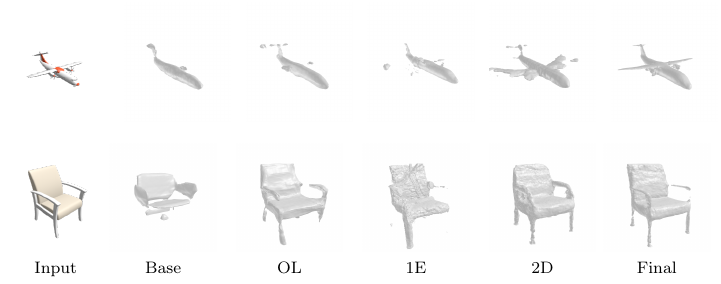}
\caption{Qualitative results obtained from the ablation study using different
network settings.}
\label{fig:qualitative_results_ablation}
\end{figure}


\setlength\tabcolsep{1.5mm}
\begin{table}
\begin{tabularx}{\columnwidth}{c|cccccc} \hline
                  & Base   & OL    & $1$E  & $2$D  & EC    &  Final         \\ \hline 
CD$\downarrow$    & 11.35  & 9.64  & 10.72 & 8.48  & 7.89  & \textbf{7.32}  \\
IoU$\uparrow$     & 51.34  & 53.95 & 51.40 & 55.23 & 55.10 & \textbf{56.83} \\
F-score$\uparrow$ & 43.11  & 48.06 & 45.92 & 51.37 & 51.33 & \textbf{52.75} \\ \hline
\end{tabularx}
\caption{Quantitative results obtained from the ablation study using different
network settings.}
\label{tab:ablation_results}
\end{table}

\subsubsection{Setup}
To investigate the impact of each individual component in our single-view 3D
reconstruction model, we performed an ablation study with the following network
options.
\begin{itemize}
  \setlength{\itemsep}{-\parsep}\setlength{\topsep}{-\parsep}
  \item \textit{Base:} A version of \modelname~that predicts the signed distance
  utilizing only global image features and coarse predictions.
  \item \textit{OL:} An improved \textit{Base} version that uses the
  probabilistic occupancy from the coarse prediction and occupancy loss.
  \item \textit{1E:} A version of \modelname~where local and global image
  features from the same encoder are used for both coarse prediction and
  localized query feature extraction.
  \item \textit{2D:} \modelname~with two separate decoders to estimate the
  signed distance from local and global query features. The final prediction is
  obtained by adding both estimations.
  \item \textit{EC:} We train \modelname~without the localization module and
  use a separate pose estimation module similar to \cite{li2021d2im} to
  predict the camera parameters. The estimated camera parameters were used to
  transform the query points during inference.
\end{itemize}
To maximize limited computational resources, we focused on the most diverse five
sub-classes (chair, car, plane, sofa, table) of the ShapeNet dataset for this
ablation study. The qualitative and quantitative results of the experiments are
recorded in Fig.~\ref{fig:qualitative_results_ablation} and
Table~\ref{tab:ablation_results} respectively.

\subsubsection{Discussion}
In the ablation experiments the \textit{Base} version was able to recover global
topology, but it lacked local geometry. As shown in
Fig~\ref{fig:qualitative_results_ablation}, the probabilistic occupancy and
optimization loss helped recover some details in the \textit{OL} version.
Conversely, the performance decreased slightly after the inclusion of local
details in the single-encoder version (\textit{1E}). We hypothesize that the
task of query point localization, while estimating the coarse prediction,
overloads the encoder and hinders meaningful feature extraction for the signed
distance prediction. To overcome this issue, we used a separate encoder for the
coarse prediction and query point localization. The dual-decoder version
(\textit{2D}), performed similar to the final model. Nonetheless, we found that
the geometric details had a thicker reconstruction than the target during
qualitative evaluation. This motivated the fusion of features rather than
predictions in the final version.

We also ablated the localization module using estimated camera parameters during
training and inference. As shown in Table~\ref{tab:ablation_results}, the final
version of \modelname~outscores the version employing estimated camera
(\textit{EC}) parameters. This indicates that our localization module with an
SDF prediction objective is more suitable for single-view reconstruction
compared to a camera pose estimation sub-module. More importantly, this removes
the requirement for pixel-wise alignment through camera parameters for local
feature extraction. Note that the \textit{EC} reconstruction appears
qualitatively similar to the others and was therefore omitted in
Fig.~\ref{fig:qualitative_results_ablation}.

\subsection{Limitations and Future Directions}
\label{subsec:limitations_and_future_directions}
Although \modelname~achieves state-of-the-art performance on single-view 3D
reconstruction, there are some limitations. For example, the model may struggle
with very small structures. We speculate that this is due to the coarse
predictor failing to provide a good estimation of such structures. Please see
the supplementary material for examples of failed reconstruction results.
Another shortcoming is the need for a clear image background. \modelname~can
reconstruct targets from real-world images, yet it requires an uncluttered
background to do this. In the future, we will work towards resolving these
issues.

\section{Conclusion}
\label{sec:conclusion}
In this paper we introduced \modelname, a network that implicitly learns how to
reconstruct a 3D object from a single image. 
Our approach does not assume weak perspective projection, nor does it require
pose estimation or rendering data. We achieved state-of-the-art performance on
single-view reconstruction from renderings of synthetic objects. Furthermore, we
demonstrated domain transferability of our model by recovering 3D surfaces from
images of real-world objects. We believe our approach could be beneficial for
other problems such as object pose estimation and novel view synthesis. 

\section*{Acknowledgments}
The authors acknowledge the Texas Advanced Computing Center (TACC) at The
University of Texas at Austin for providing software, computational, and storage
resources that have contributed to the research results reported within this
paper.

{\small
\bibliographystyle{ieee_fullname}
\bibliography{list_learning_implicitly_from_spatial_transformers_for_single-view_3d_reconstruction}

\begin{thebibliography}{10}\itemsep=-1pt

\bibitem{cao2022jiff}
Yukang Cao, Guanying Chen, Kai Han, Wenqi Yang, and Kwan-Yee~K Wong.
\newblock Jiff: Jointly-aligned implicit face function for high quality single
  view clothed human reconstruction.
\newblock In {\em Proceedings of the IEEE/CVF Conference on Computer Vision and
  Pattern Recognition}, pages 2729--2739, 2022.

\bibitem{chang2015shapenet}
Angel~X Chang, Thomas Funkhouser, Leonidas Guibas, Pat Hanrahan, Qixing Huang,
  Zimo Li, Silvio Savarese, Manolis Savva, Shuran Song, Hao Su, Jianxiong Xiao,
  Li Yi, and Fisher Yu.
\newblock Shapenet: An information-rich 3d model repository.
\newblock {\em arXiv preprint arXiv:1512.03012}, 2015.

\bibitem{chen2019learning}
Zhiqin Chen and Hao Zhang.
\newblock Learning implicit fields for generative shape modeling.
\newblock In {\em Proceedings of the IEEE/CVF Conference on Computer Vision and
  Pattern Recognition}, pages 5939--5948, 2019.

\bibitem{choy20163d}
Christopher~B Choy, Danfei Xu, JunYoung Gwak, Kevin Chen, and Silvio Savarese.
\newblock 3d-r2n2: A unified approach for single and multi-view 3d object
  reconstruction.
\newblock In {\em Proceedings of the European Conference on Computer Vision},
  pages 628--644. Springer, 2016.

\bibitem{dai2017shape}
Angela Dai, Charles Ruizhongtai~Qi, and Matthias Nie{\ss}ner.
\newblock Shape completion using 3d-encoder-predictor cnns and shape synthesis.
\newblock In {\em Proceedings of the IEEE/CVF Conference on Computer Vision and
  Pattern Recognition}, pages 5868--5877, 2017.

\bibitem{duggal2022topologically}
Shivam Duggal and Deepak Pathak.
\newblock Topologically-aware deformation fields for single-view 3d
  reconstruction.
\newblock In {\em Proceedings of the IEEE/CVF Conference on Computer Vision and
  Pattern Recognition}, pages 1536--1546, 2022.

\bibitem{fu2021single}
Kui Fu, Jiansheng Peng, Qiwen He, and Hanxiao Zhang.
\newblock Single image 3d object reconstruction based on deep learning: A
  review.
\newblock {\em Multimedia Tools and Applications}, 80(1):463--498, 2021.

\bibitem{fuentes2015visual}
Jorge Fuentes-Pacheco, Jos{\'e} Ruiz-Ascencio, and Juan~Manuel
  Rend{\'o}n-Mancha.
\newblock Visual simultaneous localization and mapping: a survey.
\newblock {\em Artificial Intelligence Review}, 43(1):55--81, 2015.

\bibitem{he2016deep}
Kaiming He, Xiangyu Zhang, Shaoqing Ren, and Jian Sun.
\newblock Deep residual learning for image recognition.
\newblock In {\em Proceedings of the IEEE/CVF Conference on Computer Vision and
  Pattern Recognition}, pages 770--778, 2016.

\bibitem{he2020geo}
Tong He, John Collomosse, Hailin Jin, and Stefano Soatto.
\newblock Geo-pifu: Geometry and pixel aligned implicit functions for
  single-view human reconstruction.
\newblock In {\em Proceedings of the Advances in Neural Information Processing
  Systems}, volume~33, pages 9276--9287, 2020.

\bibitem{jaderberg2015spatial}
Max Jaderberg, Karen Simonyan, Andrew Zisserman, and Koray Kavukcuoglu.
\newblock Spatial transformer networks.
\newblock In {\em Proceedings of the Advances in Neural Information Processing
  Systems}, volume~28, 2015.

\bibitem{kingma2014adam}
Diederik~P Kingma and Jimmy Ba.
\newblock Adam: A method for stochastic optimization.
\newblock {\em arXiv preprint arXiv:1412.6980}, 2014.

\bibitem{kruppa1913ermittlung}
Erwin Kruppa.
\newblock {\em Zur Ermittlung eines Objektes aus zwei Perspektiven mit innerer
  Orientierung}.
\newblock H{\"o}lder, 1913.

\bibitem{li2021d2im}
Manyi Li and Hao Zhang.
\newblock D2im-net: Learning detail disentangled implicit fields from single
  images.
\newblock In {\em Proceedings of the IEEE/CVF Conference on Computer Vision and
  Pattern Recognition}, pages 10246--10255, 2021.

\bibitem{list}
\url{https://github.com/robotic-vision-lab/Learning-Implicitly-From-Spatial-Transformers-Network}.

\bibitem{littwin2019deep}
Gidi Littwin and Lior Wolf.
\newblock Deep meta functionals for shape representation.
\newblock In {\em Proceedings of the IEEE/CVF International Conference on
  Computer Vision}, pages 1824--1833, 2019.

\bibitem{liu2019learning}
Shichen Liu, Shunsuke Saito, Weikai Chen, and Hao Li.
\newblock Learning to infer implicit surfaces without 3d supervision.
\newblock In {\em Proceedings of the Advances in Neural Information Processing
  Systems}, volume~32, 2019.

\bibitem{longuet1981computer}
H~Christopher Longuet-Higgins.
\newblock A computer algorithm for reconstructing a scene from two projections.
\newblock {\em Nature}, 293(5828):133--135, 1981.

\bibitem{lorensen1987marching}
William~E Lorensen and Harvey~E Cline.
\newblock Marching cubes: A high resolution 3d surface construction algorithm.
\newblock {\em ACM Siggraph Computer Graphics}, 21(4):163--169, 1987.

\bibitem{mescheder2019occupancy}
Lars Mescheder, Michael Oechsle, Michael Niemeyer, Sebastian Nowozin, and
  Andreas Geiger.
\newblock Occupancy networks: Learning 3d reconstruction in function space.
\newblock In {\em Proceedings of the IEEE/CVF Conference on Computer Vision and
  Pattern Recognition}, pages 4460--4470, 2019.

\bibitem{mittal2022autosdf}
Paritosh Mittal, Yen-Chi Cheng, Maneesh Singh, and Shubham Tulsiani.
\newblock Autosdf: Shape priors for 3d completion, reconstruction and
  generation.
\newblock In {\em Proceedings of the IEEE/CVF Conference on Computer Vision and
  Pattern Recognition}, pages 306--315, 2022.

\bibitem{nie2020total3dunderstanding}
Yinyu Nie, Xiaoguang Han, Shihui Guo, Yujian Zheng, Jian Chang, and Jian~Jun
  Zhang.
\newblock Total3dunderstanding: Joint layout, object pose and mesh
  reconstruction for indoor scenes from a single image.
\newblock In {\em Proceedings of the IEEE/CVF Conference on Computer Vision and
  Pattern Recognition}, pages 55--64, 2020.

\bibitem{niemeyer2020differentiable}
Michael Niemeyer, Lars Mescheder, Michael Oechsle, and Andreas Geiger.
\newblock Differentiable volumetric rendering: Learning implicit 3d
  representations without 3d supervision.
\newblock In {\em Proceedings of the IEEE/CVF Conference on Computer Vision and
  Pattern Recognition}, pages 3504--3515, 2020.

\bibitem{pan2019deep}
Junyi Pan, Xiaoguang Han, Weikai Chen, Jiapeng Tang, and Kui Jia.
\newblock Deep mesh reconstruction from single rgb images via topology
  modification networks.
\newblock In {\em Proceedings of the IEEE/CVF International Conference on
  Computer Vision}, pages 9964--9973, 2019.

\bibitem{park2019deepsdf}
Jeong~Joon Park, Peter Florence, Julian Straub, Richard Newcombe, and Steven
  Lovegrove.
\newblock Deepsdf: Learning continuous signed distance functions for shape
  representation.
\newblock In {\em Proceedings of the IEEE/CVF Conference on Computer Vision and
  Pattern Recognition}, pages 165--174, 2019.

\bibitem{paszke2019pytorch}
Adam Paszke, Sam Gross, Francisco Massa, Adam Lerer, James Bradbury, Gregory
  Chanan, Trevor Killeen, Zeming Lin, Natalia Gimelshein, Luca Antiga, Alban
  Desmaison, Andreas Kopf, Edward Yang, Zachary DeVito, Martin Raison, Alykhan
  Tejani, Sasank Chilamkurthy, Benoit Steiner, Lu Fang, Junjie Bai, and Soumith
  Chintala.
\newblock Pytorch: An imperative style, high-performance deep learning library.
\newblock In {\em Proceedings of the Advances in Neural Information Processing
  Systems}, volume~32, pages 8024--8035, 2019.

\bibitem{roth1982ray}
Scott~D Roth.
\newblock Ray casting for modeling solids.
\newblock {\em Computer Graphics and Image Processing}, 18(2):109--144, 1982.

\bibitem{saito2019pifu}
Shunsuke Saito, Zeng Huang, Ryota Natsume, Shigeo Morishima, Angjoo Kanazawa,
  and Hao Li.
\newblock Pifu: Pixel-aligned implicit function for high-resolution clothed
  human digitization.
\newblock In {\em Proceedings of the IEEE/CVF International Conference on
  Computer Vision}, pages 2304--2314, 2019.

\bibitem{saito2020pifuhd}
Shunsuke Saito, Tomas Simon, Jason Saragih, and Hanbyul Joo.
\newblock Pifuhd: Multi-level pixel-aligned implicit function for
  high-resolution 3d human digitization.
\newblock In {\em Proceedings of the IEEE/CVF Conference on Computer Vision and
  Pattern Recognition}, pages 84--93, 2020.

\bibitem{saputra2018visual}
Muhamad Risqi~U Saputra, Andrew Markham, and Niki Trigoni.
\newblock Visual slam and structure from motion in dynamic environments: A
  survey.
\newblock {\em ACM Computing Surveys}, 51(2):1--36, 2018.

\bibitem{schonberger2016structure}
Johannes~L Schonberger and Jan-Michael Frahm.
\newblock Structure-from-motion revisited.
\newblock In {\em Proceedings of the IEEE/CVF Conference on Computer Vision and
  Pattern Recognition}, pages 4104--4113, 2016.

\bibitem{shu20193d}
Dong~Wook Shu, Sung~Woo Park, and Junseok Kwon.
\newblock 3d point cloud generative adversarial network based on tree
  structured graph convolutions.
\newblock In {\em Proceedings of the IEEE/CVF International Conference on
  Computer Vision}, pages 3859--3868, 2019.

\bibitem{sitzmann2019deepvoxels}
Vincent Sitzmann, Justus Thies, Felix Heide, Matthias Nie{\ss}ner, Gordon
  Wetzstein, and Michael Zollhofer.
\newblock Deepvoxels: Learning persistent 3d feature embeddings.
\newblock In {\em Proceedings of the IEEE/CVF Conference on Computer Vision and
  Pattern Recognition}, pages 2437--2446, 2019.

\bibitem{sun2018pix3d}
Xingyuan Sun, Jiajun Wu, Xiuming Zhang, Zhoutong Zhang, Chengkai Zhang, Tianfan
  Xue, Joshua~B Tenenbaum, and William~T Freeman.
\newblock Pix3d: Dataset and methods for single-image 3d shape modeling.
\newblock In {\em Proceedings of the IEEE/CVF Conference on Computer Vision and
  Pattern Recognition}, pages 2974--2983, 2018.

\bibitem{tatarchenko2019single}
Maxim Tatarchenko, Stephan~R Richter, Ren{\'e} Ranftl, Zhuwen Li, Vladlen
  Koltun, and Thomas Brox.
\newblock What do single-view 3d reconstruction networks learn?
\newblock In {\em Proceedings of the IEEE/CVF Conference on Computer Vision and
  Pattern Recognition}, pages 3405--3414, 2019.

\bibitem{ullman1979interpretation}
Shimon Ullman.
\newblock The interpretation of structure from motion.
\newblock {\em Proceedings of the Royal Society of London. Series B. Biological
  Sciences}, 203(1153):405--426, 1979.

\bibitem{wu2020pq}
Rundi Wu, Yixin Zhuang, Kai Xu, Hao Zhang, and Baoquan Chen.
\newblock Pq-net: A generative part seq2seq network for 3d shapes.
\newblock In {\em Proceedings of the IEEE/CVF Conference on Computer Vision and
  Pattern Recognition}, pages 829--838, 2020.

\bibitem{xu2019disn}
Qiangeng Xu, Weiyue Wang, Duygu Ceylan, Radomir Mech, and Ulrich Neumann.
\newblock Disn: Deep implicit surface network for high-quality single-view 3d
  reconstruction.
\newblock In {\em Proceedings of the Advances in Neural Information Processing
  Systems}, volume~32, 2019.

\bibitem{zhang2021holistic}
Cheng Zhang, Zhaopeng Cui, Yinda Zhang, Bing Zeng, Marc Pollefeys, and
  Shuaicheng Liu.
\newblock Holistic 3d scene understanding from a single image with implicit
  representation.
\newblock In {\em Proceedings of the IEEE/CVF Conference on Computer Vision and
  Pattern Recognition}, pages 8833--8842, 2021.

\bibitem{zhao2021learning}
Fang Zhao, Wenhao Wang, Shengcai Liao, and Ling Shao.
\newblock Learning anchored unsigned distance functions with gradient direction
  alignment for single-view garment reconstruction.
\newblock In {\em Proceedings of the IEEE/CVF International Conference on
  Computer Vision}, pages 12674--12683, 2021.

\end{thebibliography}
}

\setcounter{section}{0}
\section*{Supplementary Material}
\begin{figure}[ht]
\centering
\includegraphics[trim={30 0 30 0},clip,width=\columnwidth]{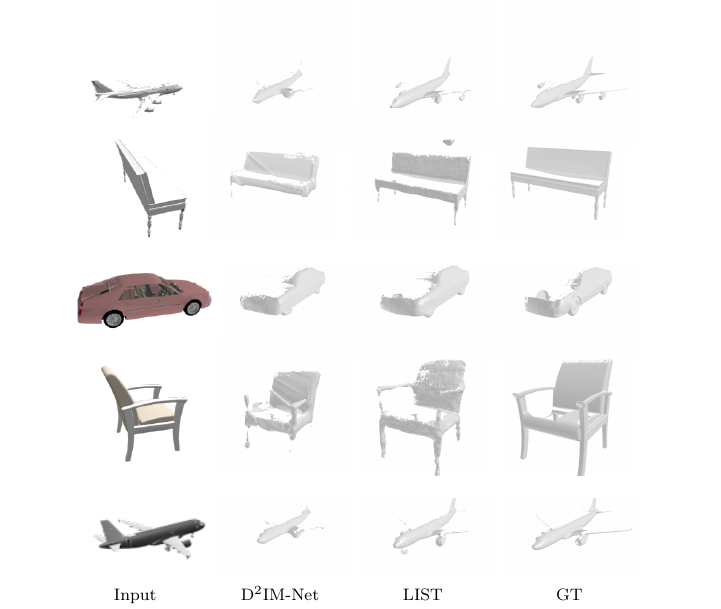}
\caption{A qualitative comparison between \modelname~and the baseline models on
occluded surface reconstruction using the ShapeNet dataset. GT denotes the
ground-truth objects.}
\label{fig:qualitative_results_occluded}
\end{figure}

\begin{figure}[ht]
\centering
\includegraphics[width=\columnwidth]{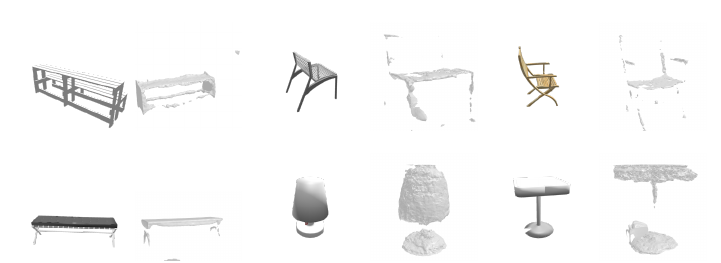}
\caption{Examples of failed \modelname~reconstructions.}
\label{fig:failed_results}
\end{figure}

\begin{figure*}[ht]
\centering
\includegraphics[trim={30 0 30 0},clip,width=\textwidth,height=0.92\textheight]{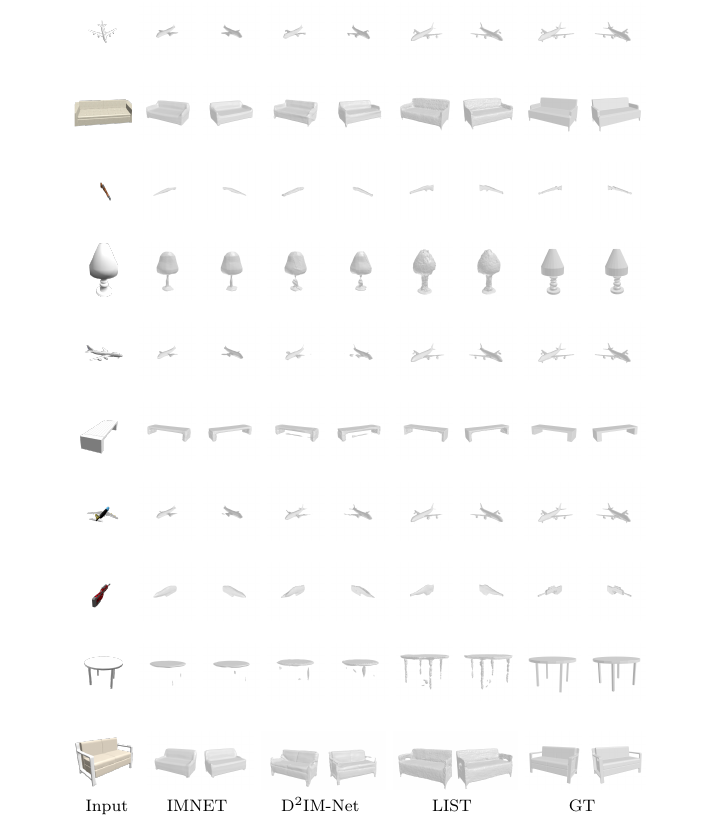}
\caption{A qualitative comparison between \modelname~and the baseline models
using the ShapeNet dataset. Our model recovers \textit{significantly better}
topological and geometric structure, and the reconstruction is not tainted by
the input-view direction. GT denotes the ground-truth objects.}
\label{fig:qualitative_results_comarison_1}
\end{figure*}

\begin{figure*}[ht]
\centering
\includegraphics[width=0.99\textwidth]{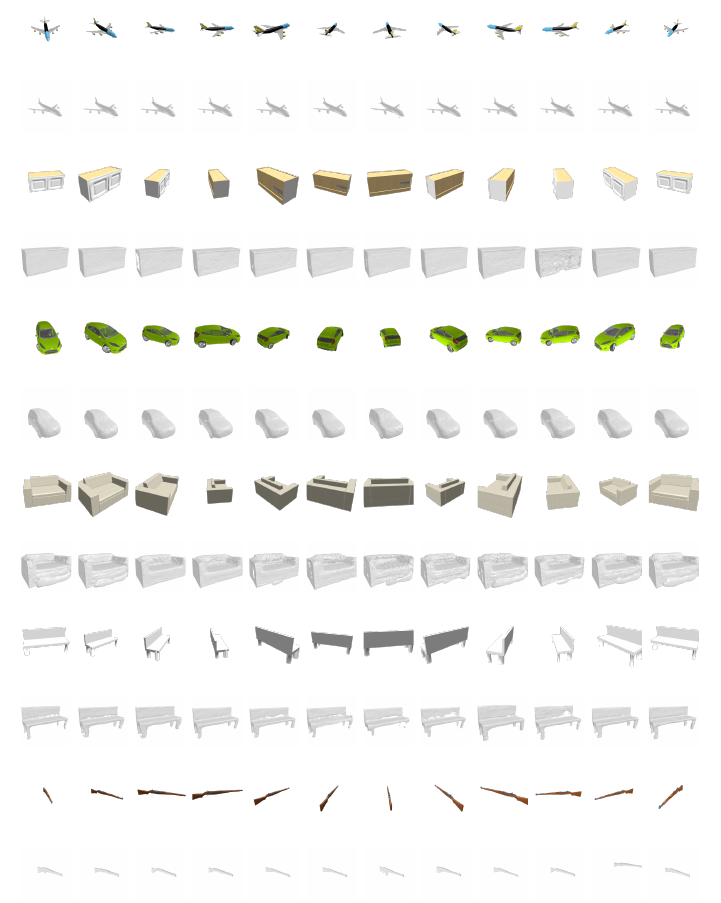}
\caption{Qualitative results of \modelname~reconstructions using distinct views
of the same object. Odd rows represent the input and even rows represent the
reconstructions.} 
\label{fig:qualitative_results_view_1}
\end{figure*}

\begin{figure*}[ht]
\centering
\includegraphics[width=0.99\textwidth]{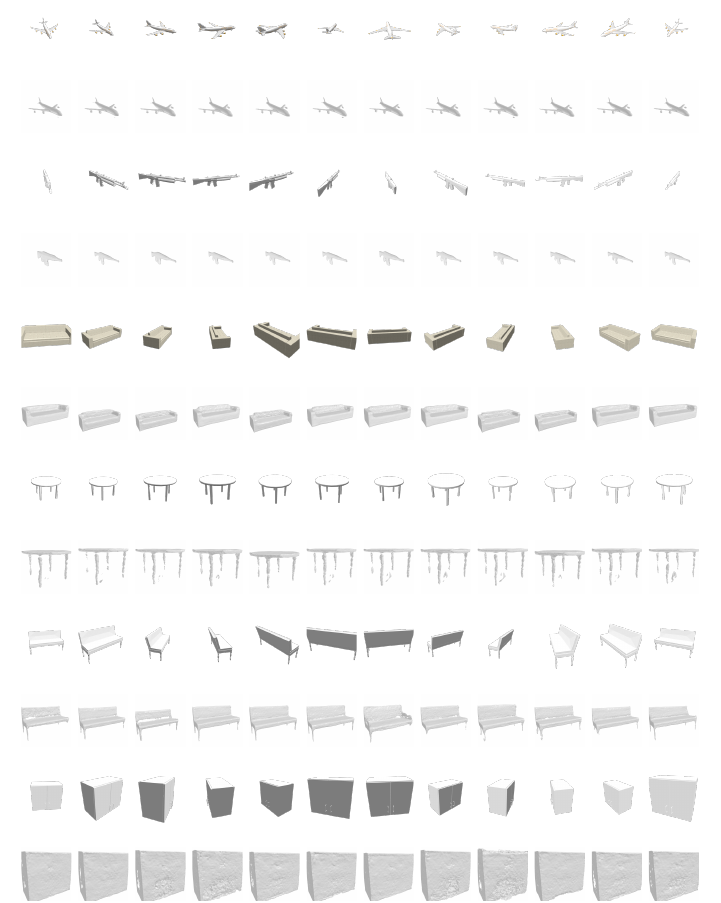}
\caption{Qualitative results of \modelname~reconstructions using distinct views
of the same object. Odd rows represent the input and even rows represent the
reconstructions.} 
\label{fig:qualitative_results_view_2}
\end{figure*}

\begin{figure*}[ht]
\centering
\includegraphics[trim={0 25 0 0},clip,width=0.99\textwidth]{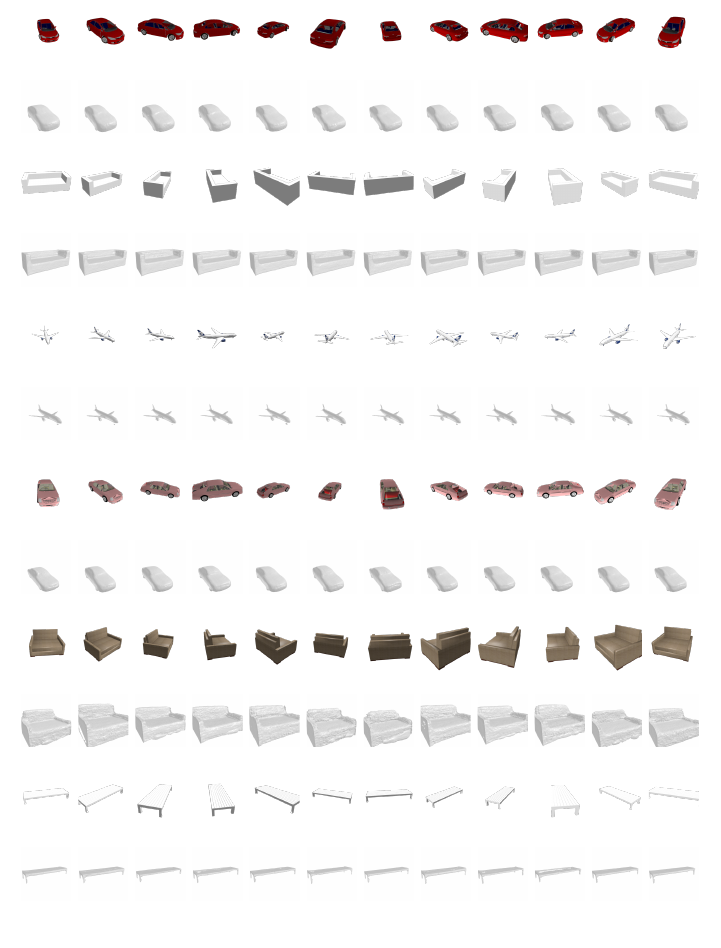}
\caption{Qualitative results of \modelname~reconstructions using distinct views
of the same object. Odd rows represent the input and even rows represent the
reconstructions.} 
\label{fig:qualitative_results_view_3}
\end{figure*}

In Fig.~\ref{fig:qualitative_results_occluded}, we show a
qualitative comparison of occluded surface reconstruction. Examples of failed
reconstructions are displayed in Fig.~\ref{fig:failed_results}. More qualitative
comparisons between \modelname~and the baseline models using the ShapeNet
dataset are highlighted in Fig.~\ref{fig:qualitative_results_comarison_1}.
The results of \modelname~reconstructions using distinct views of the same object
are provided in Fig.~\ref{fig:qualitative_results_view_1},
Fig.~\ref{fig:qualitative_results_view_2}, and
Fig.~\ref{fig:qualitative_results_view_3}. 

\section{Evaluation Metrics} 
\textbf{Chamfer Distance (CD)}: 
The chamfer distance (CD) between two meshes is defined as
\begin{equation}
  \text{CD}(y_{\text{GT}}, y_{\text{pred}}) = \sum_{a\in y_{pred}}^{}\min_{b \in y_{gt}} {||a-b||} + \sum_{b\in y_{gt}}^{}\min_{b\in y_{pred}} {||b-a||},
\end{equation}
where, $y_{GT}$ and $y_{pred}$ are two point clouds extracted from the surface
of the ground-truth and reconstructed object, respectively.

\textbf{Intersection over Union (IoU)}: 
The volumetric intersection over union (IoU) is defined as the quotient of the
volume of the intersection of two meshes and the volume of their union,
\begin{equation}
  \text{IoU}(\mathcal{M}_{\text{pred}}, \mathcal{M}_{\text{GT}}) = \frac{|\mathcal{M}_{\text{pred}} \cap \mathcal{M}_{\text{GT}}|}{|\mathcal{M}_{\text{pred}} \cup \mathcal{M}_{\text{GT}}|}.
\end{equation}

\textbf{F-score}:
The F-score, proposed in \cite{tatarchenko2019single} as a comprehensive scoring
metric for single-view reconstruction, combines precision and recall to quantify
the overall reconstruction quality. Concretely, the F-score at a distance
threshold $d$ is given by
\begin{equation*}
  F(d) = \frac{2 \cdot P(d) \cdot R(d)}{P(d) + R(d)},
\end{equation*} 
where $P(\cdot)$ and $R(\cdot)$ represents the precision and recall,
respectively. Precision quantifies the accuracy while recall assesses the
completeness of the reconstruction. For the ground-truth $y_{gt}$ and
reconstructed point cloud $y_{pred}$, the precision of an outcome at $d$ can be
calculated as
\begin{equation*}
  P(d) = \sum_{i \in y_{\text{pred}}}^{}[\min_{j \in y_{\text{GT}}} ||i-j|| < d].
\end{equation*}
Similarly, the recall for a given $d$ may be computed as
\begin{equation*}
  R(d) = \sum_{j \in y_{\text{GT}}}^{}[\min_{i \in y_{\text{pred}}} ||j-i|| < d].
\end{equation*}
To evaluate the reconstructions between \modelname~and the baselines we used
$d=1\%$.

\section{Data Preparation} 
To prepare the ground truth, first the target shape was normalized into a unit
cube and 50k points were sampled from the surface of the object. The query
points were prepared by adding random Gaussian noise ($n$) to the surface
points. Specifically,
\begin{equation}
  Q_j = Q_S + n~|~n \in \mathcal{N}(0, P),
\end{equation}
where $Q_S$ are the sampled points and $P \in \mathbb{R}^{3 \times 3}$ is a
diagonal covariance matrix with entries $P_{i,i} = \rho$. We empirically found
that $45\%$ of the points at $\rho = 0.003$, $44\%$ of the points at $\rho =
0.01$, and $10\%$ of the points at $\rho = 0.07$ achieved the best results.

\section{Implementation, Training, and Inference Details} 
\subsection{Implementation Overview}
\modelname~was implemented using the PyTorch \cite{paszke2019pytorch} library.
To optimize the model, the Adam \cite{kingma2014adam} optimizer was used with
coefficients $(0.9,0.99)$, learning rate $10^{-4}$, and weight decay $10^{-5}$.
A pretrained ResNet \cite{he2016deep} was employed as the image encoder in
$\Omega$ and $\Pi$. We closely followed the generator in \cite{shu20193d} to
implement the coarse predictor in $\Omega$ with tree-structured convolutions.
However, we empirically found that the degree values $(2,2,2,2,2,2,64)$ provided
a better coarse estimation in our settings. We set the coarse point cloud
density to $N=4000$, and the occupancy grid resolution to $M=128$. To generate a
probabilistic occupancy with the same grid, we utilized a shallow convolutional
network $\Gamma$.

We define $\Xi$ as a convolutional neural network to map the probabilistic
occupancy grid into a high-dimensional latent space. To extract the global query
features and localize the query points, we used a fully-connected neural network
$\Theta$. The global image features are fused with the global query features on
the 3rd layer of $\Theta$. During training, we augment the images with random
color jitter, and normalize the values to $[0,1]$. To improve the estimation
accuracy, we scale the ground-truth and predicted SDF values by $10.0$.
Following \cite{chen2019learning}, we disentangled the query points by scaling
with $2.0$ and swapping the 1st and 3rd axis to extract query features from the
coarse prediction. At test time, we extract the query points from a grid in the
range $[-0.5, 0.5]$ with resolution $128^3$.

\section{Training and Inference Time} 
To train \modelname~it takes $\approx$ 1 s to make a forward pass on an Intel
i7 machine with an NVIDIA GeForce GTX 1080Ti GPU. To fully pass through the
Pix3D and ShapeNet datasets, it takes approximately 35 and 50 min,
respectively. Our training process involved using 4 1080Ti GPUs for 100 epochs
with a batch size of 8. To reconstruct the mesh of a single object from a
corresponding RGB image, it takes $\approx$ 7 s on average at a grid resolution
of $128^3$.

\end{document}